\documentclass[10pt, a4paper]{article}

\usepackage{lrec-coling2024} 
\usepackage{booktabs}
\usepackage{tablefootnote}
\usepackage{multirow}
\usepackage{float}

\newif{\ifhidecomments}
\hidecommentsfalse

\ifhidecomments
    \newcommand{\botond}[1]{}
    \newcommand{\attila}[1]{}
    \newcommand{\judit}[1]{}
    \newcommand{\dorina}[1]{}
\else
    \newcommand{\botond}[1]{\textcolor{red}{[#1 ({\bf Botond})]}}
    \newcommand{\attila}[1]{\textcolor{blue}{[#1 ({\bf Attila})]}} 
    \newcommand{\judit}[1]{\textcolor{orange}{[#1 ({\bf Judit})]}} 
    \newcommand{\dorina}[1]{\textcolor{green}{[#1 ({\bf Dorina})]}}
\fi

\title{\textbf{From News to Summaries: Building a Hungarian Corpus for Extractive and Abstractive Summarization}}

\name{Botond Barta$^{1,2}$, Dorina Lakatos$^{1,2}$, Attila Nagy$^{2}$, Mil\'an Konor Nyist$^{2}$, Judit \'Acs$^{1}$}

\address{$^{1}$HUN-REN Institute for Computer Science and Control\\
        $^{2}$Department of Automation and Applied Informatics\\
        Budapest University of Technology and Economics\\
         botondbarta@sztaki.hu, dorinapetra@gmail.com, attila.nagy234@gmail.com \\ nyist.milan78@gmail.com, acsjudit@sztaki.hu\\}

\abstract{
Training summarization models requires substantial amounts of training data. However for less resourceful languages like Hungarian, openly available models and datasets are notably scarce. To address this gap our paper introduces HunSum-2 an open-source Hungarian corpus suitable for training abstractive and extractive summarization models. The dataset is assembled from segments of the Common Crawl corpus undergoing thorough cleaning, preprocessing and deduplication. In addition to abstractive summarization we generate sentence-level labels for extractive summarization using sentence similarity. We train baseline models for both extractive and abstractive summarization using the collected dataset. To demonstrate the effectiveness of the trained models, we perform both quantitative and qualitative evaluation. Our dataset, models and code are publicly available, encouraging replication, further research, and real-world applications across various domains.
 \\ \newline \Keywords{abstractive summarization, extractive summarization, Hungarian}  }

\begin{document}

\maketitleabstract

\section{Introduction}
The goal of Automatic Text Summarization is to produce a short, concise text, which retains key information from a longer article \citep{mani1999advances}. The advent of pre-trained language models has significantly advanced the field with a large body of research now concentrated on leveraging these models for more effective and coherent summaries \citep{liu-lapata-2019-text}. The two main approaches to summarization are extractive and abstractive. 

Extractive summarization methods identify and extract salient sentences or tokens directly from the source document to construct the summary \citep{cao-etal-2016-attsum, cheng-lapata-2016-neural}. These models are generally less coherent, but faster and less prone to faithfulness related problems compared to their abstractive counterpart \citep{li-etal-2021-ease, dreyer-etal-2023-evaluating}. In recent years, pre-trained language models such as GPT \citep{brown2020language}, PEGASUS \citep{zhang2020pegasus} and T5 \citep{raffel2020exploring} have shown promising results in generating abstractive summaries. Although these models produce very fluent summaries, they tend to hallucinate inconsistent or contradictory content compared to the source document \citep{maynez-etal-2020-faithfulness}.

In this paper, we build a dataset for Hungarian summarization and release it as open-source\footnote{\url{https://github.com/botondbarta/HunSum}} alongside models trained on the data. We construct an abstractive summarization corpus\footnote{\href{https://huggingface.co/datasets/SZTAKI-HLT/HunSum-2-abstractive}{SZTAKI-HLT/HunSum-2-abstractive}} by performing a thorough cleaning and preprocessing of Hungarian segments from the Common Crawl dataset. Using the crawled news articles we also generate an extractive summarization corpus\footnote{\href{https://huggingface.co/datasets/SZTAKI-HLT/HunSum-2-extractive}{SZTAKI-HLT/HunSum-2-extractive}} by selecting the most similar article sentence for each lead sentence based on their sentence embeddings. We train both abstractive and extractive models on this corpus and evaluate them both quantitatively and qualitatively.

\section{Related work}
The CNN-DM corpus \citep{nallapati-etal-2016-abstractive} was the first large-scale English abstractive summarization dataset which was constructed by scraping news outlets. Their summaries used human-generated summary bullets on the page. Another English-language summarization dataset is XSum \citep{narayan-etal-2018-dont} which uses specific HTML classes on the page to collect the summary. Several different monolingual datasets have been inspired by XSum such as the French OrangeSum \citep{kamal-eddine-etal-2021-barthez} or the Russian Gazeta \citep{gusev2020dataset}. We follow a similar methodology later on in our paper.  For Hungarian summarization \citet{yang} build a corpus from two major Hungarian news sites (overlapping with our dataset) and train BERT-like models \citep{devlin-etal-2019-bert}. \citet{presumm_yang} train multilingual and Hungarian models based on PreSumm \citep{presumm}. \citet{makraitowards} train an encoder-decoder model based on huBERT \citep{Nemeskey:2020} using the ELTE.DH corpus \cite{indig-elte-dh}. \citet{barterezzunk} train BART-based models \citep{lewis-etal-2020-bart} for abstractive summarization. \citet{yang-multi-sum} fine-tune PEGASUS and multilingual models mT5 and mBART for Hungarian abstracive summarization. We do our best effort to compare models trained on our dataset to prior works. Most works in Hungarian only released models and not the datasets, so any comparative analysis has to be taken with a grain of salt. A prior version of this dataset was released as HunSum-1 \citep{hunsum1} with less preprocessing, fewer data sources and no extractive summaries.

\section{Methods}

\subsection{Dataset collection}
We use the freely available Common Crawl dataset\footnote{https://commoncrawl.org/} as a basis for constructing the corpus. It contains petabytes of crawled web pages from the past 25 years and it is available on Amazon S3 in WARC format. Retrieval and deduplication of the raw dataset by domains was done using the downloader created by \citet{Nemeskey:2020}. We pick 27 Hungarian news sites including most major Hungarian-language news sites to build our corpus. The selected sites all have a dedicated lead article field to make extracting the summary easier. The final raw dataset was 290 GB of data in HTML format. We then extracted the relevant parts from each article: the lead, the article, the title, the creation date and optionally some tags. We apply the following preprocessing steps and constraints:
\begin{itemize}
    \item Remove links, image captions and embedded social media from articles.
    \item Remove galleries.
    \item Discard articles that are a part of a live blog.
    \item Discard articles where the article text is shorter than the lead.
    \item Discard articles shorter than 200 characters or longer than 15,000 characters or have fewer than 6 sentences.
    \item Discard articles with leads shorter than 6 tokens or longer than 5 sentences.
    \item Remove low-quality or incorrectly scraped data points. We assess quality by calculating the similarity between the leads and articles using the \verb|paraphrase-multilingual-MiniLM-L12-v2| from the \verb|sentence-transformer| package and remove those with a similarity score below 0.17.
\end{itemize}
Through exploratory data analysis we also removed problematic patterns in the data, such as lottery and sports results, where the data was not applicable to summarization.

\begin{figure}[!htbp]
    \includegraphics[width=0.99\linewidth]
    {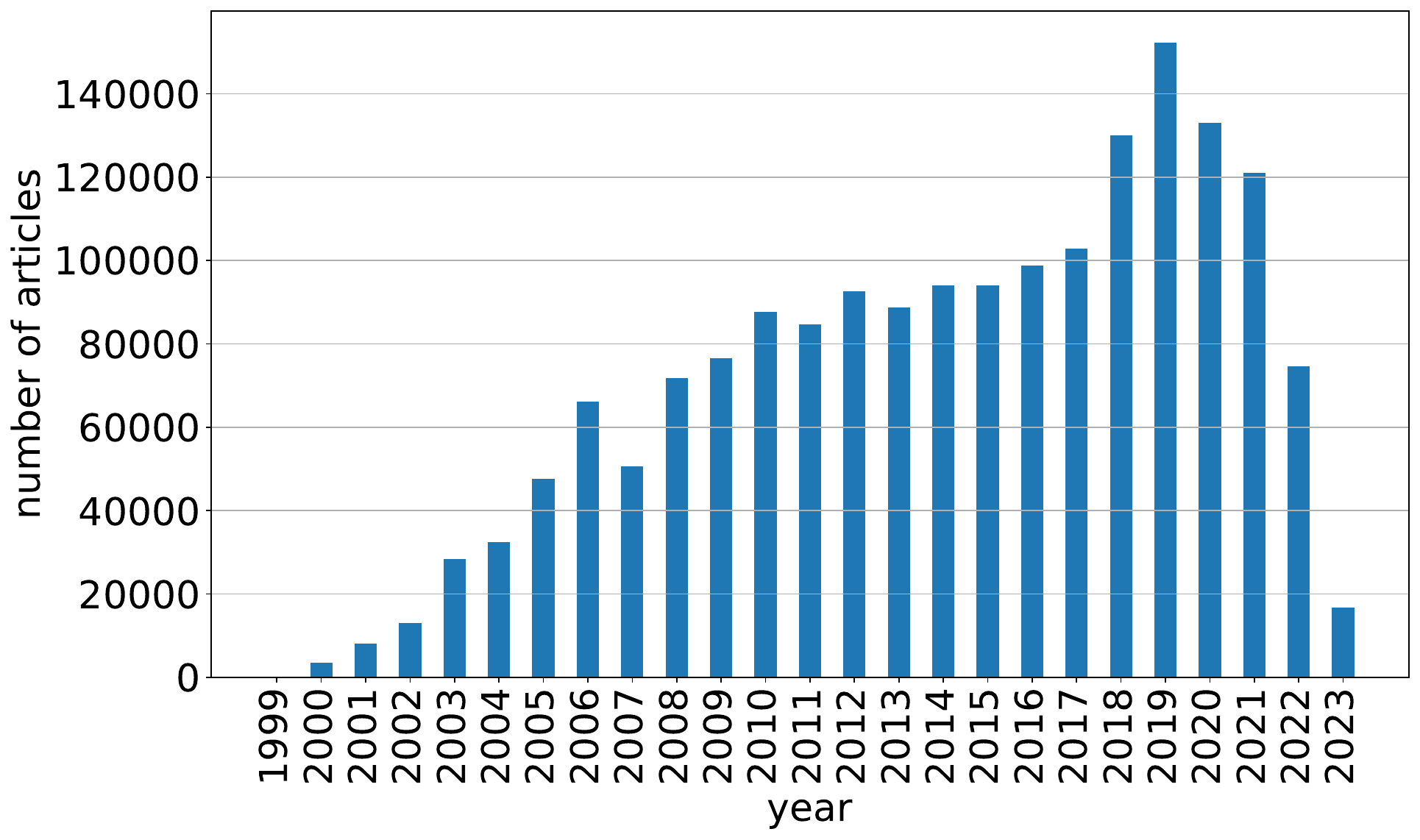}
    \caption{Number of articles by year.}
    \label{fig:article-by-year}
\end{figure}

For tokenization and sentence splitting, we used the quntoken\footnote{https://github.com/nytud/quntoken} package, for language detection we used FastText \citep{joulin-etal-2017-bag}. We also remove near-duplicate documents with Locality Sensitive Hashing (LSH) with a similarity threshold of 0.45. If two articles were classified as similar, we kept the more recent one. The preprocessed and deduplicated dataset contains 1.82 million documents. Distribution by year and source with the average sentence and token numbers can be seen in Figure \ref{fig:article-by-year} and Table \ref{Tab:data_stats_per_site}. We also compute a number of commonly used descriptive statistical measures about the dataset such as Novel N-gram ratio (NNG-n) \citep{narayan-etal-2018-dont}, compression (CMP) \citep{bommasani-cardie-2020-intrinsic} and redundancy (RED-n) \citep{xlsum} listed in Table~\ref{table:automatic-metrics}.

\begin{table*}[!htbp]
  \centering
    \begin{tabular}{ c  c  c  c  c  c }
    \toprule
    \textbf{NNG-1} & \textbf{NNG-2} & \textbf{NNG-3} & \textbf{CMP} & \textbf{RED-1} & \textbf{RED-2} \\
    \midrule
    41.12 & 77.31 & 88.74 & 89.1 & 11.78 & 0.51 \\
    \bottomrule
    \end{tabular}
  \caption{Intrinsic evaluation of the dataset.}
  \label{table:automatic-metrics}
\end{table*}

We split the final dataset with stratified sampling using the news sources to train-dev-test with the dev and tests being 1998 documents. This split is released alongside the entire dataset on Huggingface. We carry out all of our experiments on this split and encourage further works to do so for comparable results.

\begin{table*}[!htbp]
\centering
  \begin{tabular}{l r r r r r r r}
    \toprule
    \textbf{} & \textbf{}  & \multicolumn{2}{c}{Article} & \multicolumn{2}{c}{Lead} \\
    \cmidrule(lr){3-4}                  
    \cmidrule(lr){5-6}
    \textbf{Site} & \textbf{Count}  & \textbf{tokens} & \textbf{sents} & \textbf{tokens} & \textbf{sents} \\
    \midrule
    \textit{regional} & 346 812 & 368.2 & 18.6 & 27.1 & 1.5 \\
24.hu & 307 477 & 350.6 & 18.8 & 22.7 & 1.4 \\
origo.hu & 293 810 & 408.3 & 20.3 & 40.5 & 2.0 \\
hvg.hu & 206 719 & 382.4 & 17.1 & 30.0 & 1.5 \\
kisalfold.hu & 161 315 & 341.6 & 18.8 & 25.8 & 1.5 \\
index.hu & 159 545 & 526.2 & 26.1 & 42.5 & 2.2 \\
delmagyar.hu & 153 139 & 351.4 & 18.9 & 29.8 & 1.7 \\
nlc.hu & 99 674 & 385.2 & 22.1 & 26.1 & 1.7 \\
nepszava.hu & 28 493 & 468.2 & 21.4 & 33.2 & 1.6 \\
portfolio.hu & 22 766 & 470.2 & 21.5 & 54.3 & 2.1 \\
m4sport.hu & 19 673 & 397.7 & 24.8 & 28.7 & 1.3 \\
metropol.hu & 12 007 & 295.7 & 15.9 & 25.1 & 1.4 \\
telex.hu & 6 420 & 918.9 & 41.6 & 52.0 & 2.4 \\
    \bottomrule
  \end{tabular}
  \caption{Average length of the articles and leads. The \textit{regional} category groups smaller, local news sites.}
  \label{Tab:data_stats_per_site}
\end{table*}

\subsection{Abstractive Summarization}
We trained baseline models using our dataset. As there is no publicly available Hungarian generative model, we experimented with mT5 \citep{mt5}, the multilingual version of the T5 model. Another model we experimented with is the Hungarian version of the BERT model, huBERT \citep{Nemeskey:2020}, which we fine-tuned as an encoder-decoder architecture (Bert2Bert).

\begin{table}[!htbp]
\centering
  \begin{tabular}{l r}
    \toprule
    \textbf{Parameter}  & \textbf{Bert2Bert/mT5} \\
    \midrule
    batch size & 13   \\
    learning rate & 5e-5 \\
    weight decay & 0.01   \\
    warmup steps & 16000/3000 \\
    patience & 6 \\
    \bottomrule
  \end{tabular}
  \caption{Hyperparameters for training abstractive summarization models.}
  \label{Tab:hyperparams}
\end{table}


We fine-tuned these models on our dataset using the parameters in Table \ref{Tab:hyperparams}. The BERT models have a maximum input length of 512 tokens, and for comparison purposes we also truncated the input in case of the mT5 model. The models were trained on a single NVIDIA A100 GPU with early stopping on the validation loss. The mT5 model stopped learning at 8.14 epoch, while the Bert2Bert model at 3.8.

\subsection{Extractive Summarization}

Extractive summarization models highlight sentences that summarize the article.
Training such models requires binary labeling at the sentence level which is not available in our raw dataset. To transform our data into this form, we used sentence transformers to calculate the embedding of the lead and article sentences, and then for each lead sentence we selected the closest article sentence by cosine distance in such a way that the sum of similarities is maximised. The sentence embeddings were computed using the \verb|paraphrase-multilingual-MiniLM-L12-v2| model.

We chose the \textsc{BertSum} \citep{liu-bertsum} architecture using huBERT with a simple classifier layer at the end to train our baseline model for extractive summarization. To train our model we used the same train-dev-test split mentioned before. The model was trained for 21,000 steps using a batch size of 200 with a learning rate of 5e-5. We evaluated the model every 1000 steps on our validation set and stopped the training process when the evaluation loss had not decreased in 10 evaluation step. The model was trained on four NVIDIA A100 GPUs.

\section{Results}

\subsection{Quantitative Evaluation}
We evaluated our abstractive and extractive models using two automatic metrics: ROUGE \cite{lin-2004-rouge} and BertScore \cite{bert_score}. The results can be seen in Table \ref{Tab:results}. The extractive model outperformed the abstractive models significantly in terms of ROUGE and slightly in terms of BertScore. This may be a biased comparison to some extent, since the extractivity of the dataset itself favors extractive models when making comparisons using metrics such as ROUGE. We also compared our models to other publicly available Hungarian abstractive summarization models. The ROUGE scores turned out considerably lower for these models with a multilingual BART model producing the highest ROUGE score. As these models' training and test data is not available, we only evaluated them on our test set, this likely explains the performance difference compared to our models. We also compared our best performing abstractive model Bert2Bert with other models trained on monolingual summarization datasets in other languages. For most of them, only ROUGE scores have been published, therefore only these are shown in Table \ref{Tab:comparing-results}. Due to the varying sizes of the other publicly available datasets and their linguistic differences, it is not possible to draw any major conclusions except that the ROUGE scores of the models are roughly in the same range.

\begin{table*}[!htbp]
\centering
\setlength{\cmidrulekern}{0.25em}
  \begin{tabular}{l r r r r }
    \toprule
    \textbf{Model} & \textbf{R-1}  & \textbf{R-2} & \textbf{R-L} & \textbf{BertScore} \\
    \midrule
    Bert2Bert  & 40.95  & 14.18  & 27.42 & 78.81  \\
    mT5-base   & 40.06 & 12.67 & 25.93 & 78.64  \\
    extractive & \textbf{49.85} & \textbf{20.12} & \textbf{33.46} & \textbf{79.18} \\
    \midrule
    hi-mbart-large-50 \cite{yang-multi-sum} & 31.63 & 13.26  & 22.82 & 77.77  \\
    hi-mt5-base \cite{yang-multi-sum} & 29.53 &  11.34  & 21.35 & 76.99  \\
    foszt2oszt \cite{makraitowards} & 26.87 &  8.03  & 20.19 & 75.84  \\
    \bottomrule
  \end{tabular}
  \caption{ROUGE and BertScore recall scores on the test set. ROUGE-1, ROUGE-2 and ROUGE-L scores are abbreviated as R-1, R-2 and R-L respectively.}
  \label{Tab:results}
\end{table*}

\begin{table*}[!htbp]
\centering
\setlength{\cmidrulekern}{0.25em}
  \begin{tabular}{l l r r r r}
    \toprule
    \textbf{Dataset} & \textbf{Language} & \textbf{Size} & \textbf{R-1}  & \textbf{R-2} & \textbf{R-L} \\
    \midrule
    HunSum-2 (ours) & Hungarian & 1.82M & 40.95  & 14.18  & 27.42 \\
    CNN/DM \citep{nallapati-etal-2016-abstractive} & English & 312K & 35.46 & 13.30 & 32.65 \\
    OrangeSum \cite{kamal-eddine-etal-2021-barthez} & French & 30K & 32.67 &  13.73  & 23.18  \\
    pn-summary \cite{farahani2021leveraging} & Persian & 93K & 44.01 &  25.07  & 37.76  \\
    Gazeta \cite{gusev2020dataset} & Russian & 60K & 32.10 &  14.20  & 27.90  \\
    IlPost \cite{landro2022two} & Italian & 44K & 38.91 &  21.38 & 32.05  \\
    \bottomrule
  \end{tabular}
  \caption{ROUGE scores on different monolingual abstractive summarization models.}
  \label{Tab:comparing-results}
\end{table*}

\subsection{Qualitative Evaluation}
Quantitative metrics cannot always reveal specific problems with abstractive summarization models, such as hallucinations or biases. For this reason, we conduct a qualitative analysis on a 60 document sample from the test set. We extend the questions used by \citet{xlsum} with an additional question about grammaticality. Each annotator has to answer the following questions for each model prediction:
\begin{itemize}
    \item \textbf{Relevant:} Does the summary convey what the article is about?
    \item \textbf{Consistent:} Does the summary only contain information that is consistent with the article?
    \item \textbf{No Hallucination:} Does the summary only contain information that can be inferred from the article?
    \item \textbf{Grammatical:} Is the summary grammatically correct?
\end{itemize}
Annotators are also asked, which summary they consider best, in that case the extractive model summary is also an option to select.

\begin{figure}[!htbp]
    \includegraphics[width=0.99\linewidth]
    {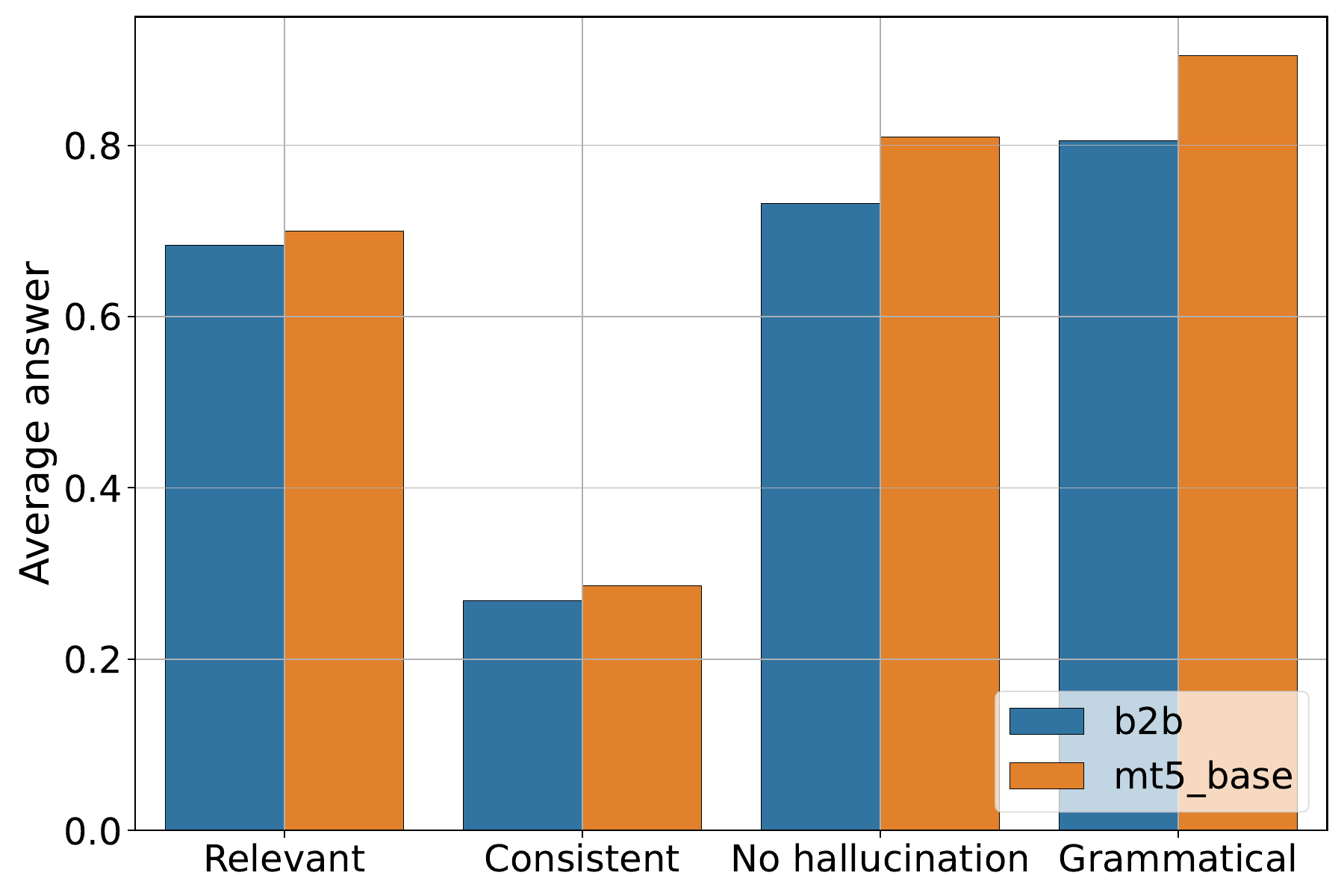}
    \caption{The average answers for the properties by models.}
    \label{fig:qualitative_eval}
\end{figure}

All annotators are native Hungarian speakers. Every data point was annotated by three annotators. The average majority answers are presented in Figure \ref{fig:qualitative_eval} where 1 means \texttt{Yes} and 0 means \texttt{No}. The average pairwise Cohen kappa between the annotators is 0.60 indicating moderate agreement. The results show that the mT5 model performs slightly better on all 4 questions. In general, close to 70\% of the articles were classified as correctly capturing the gist of the document for both models. Factuality seems to be the biggest pain point as close to two thirds of the generations contained at least one inconsistency with the original article. Interestingly outputs that cannot be verified from the source sentence (extrinsic hallucinations) were produced less frequently, only in about 20\% of cases for the mT5 model. For the question about the best model, the extractive model was chosen 60\% of the time, while the mT5 model only reached 23\%. Annotators felt that although extractive summaries were often less coherent, the factual mistakes and inconsistencies made abstractive summaries less desirable.

\section{Conclusion}
This paper presents a novel open-source Hungarian corpus designed for training both extractive and abstractive summarization models. The baseline models trained on the dataset have shown promising results both quantitatively and qualitatively with the extractive model performing best. Although the abstractive models produced fluent and grammatically correct sentences, the qualitative evaluation highlighted concerns particularly around factuality. Improving this is an exciting future direction both via making improvements to the dataset or experimenting with architectures that optimize for factual correctness. We encourage future works to use this dataset for benchmarking new methods for Hungarian summarization and hope that this will improve reproducibility in the field.

\section{Acknowledgements}
This study was supported by the European Union project RRF-2.3.1-21-2022-00004 within the framework of the Artificial Intelligence National Laboratory, Hungary.


\nocite{*}
\section{Bibliographical References}\label{sec:reference}

\bibliographystyle{lrec-coling2024-natbib}
\bibliography{lrec-coling2024-example}

\bibliographystylelanguageresource{lrec-coling2024-natbib}
\bibliographylanguageresource{languageresource}

\end{document}
